\documentclass{llncs}
\usepackage{cite}
\usepackage{graphicx}
\usepackage{epstopdf}
\graphicspath{{eps/}}
\usepackage{upgreek}
\DeclareGraphicsExtensions{.eps}
\usepackage[cmex10]{amsmath}
\usepackage{amsfonts}
\usepackage{amssymb}
\usepackage{enumerate}
\usepackage{url}
\usepackage{multirow}
\usepackage{scrextend}
\usepackage{tabularx}
\usepackage{times}
\usepackage{fixltx2e}
\usepackage{subfigure}
\usepackage{floatrow}
\newcommand{\eg}{\emph{e.g.},}
\newcommand{\etal}{\emph{et al.}}
\newcommand{\ie}{\emph{i.e.},}

\begin{document}
\title{Characterization of Lung Nodule Malignancy using Hybrid Shape and
  Appearance Features}
\titlerunning{{Characterization of Lung Nodule Malignancy with Hybrid
    Features}}
\author{Mario Buty \inst{1} \and Ziyue Xu \inst{1}
  \thanks{Corresponding author: ziyue.xu@nih.gov. This research is
    supported by CIDI, the intramural research program of the National
    Institute of Allergy and Infectious Diseases (NIAID) and the
    National Institute of Biomedical Imaging and Bioengineering
    (NIBIB).}\and Mingchen Gao \inst{1} \and Ulas Bagci \inst{2} \and
  Aaron Wu \inst{1} \and \\ Daniel J. Mollura \inst{1}}
\institute{National Institutes of Health, Bethesda, MD, USA. \\ \and University of Central Florida, Orlando, FL, USA.}
\authorrunning{M. Buty, Z. Xu, M. Gao, U. Bagci, A. Wu, and D. J. Mollura}

\maketitle
\begin{abstract}
Computed tomography imaging is a standard modality for detecting
and assessing lung cancer. In order to evaluate the malignancy of
lung nodules, clinical practice often involves expert qualitative
ratings on several criteria describing a nodule's appearance and shape. Translating these features for
computer-aided diagnostics is challenging due to their subjective nature and the difficulties in gaining
a complete description. In this paper, we propose a computerized approach to
quantitatively evaluate both appearance distinctions and 3D surface
variations. Nodule shape was modeled and parameterized using spherical harmonics, and appearance features were extracted using deep
convolutional neural networks. Both sets of features were combined to
estimate the nodule malignancy using a random forest classifier. The
proposed algorithm was tested on the publicly available Lung
Image Database Consortium dataset, achieving high accuracy. By providing lung nodule characterization,
this method can provide a robust alternative reference opinion for lung cancer diagnosis.

\end{abstract}
\section{Introduction}
\label{sec:intro}
Lung cancer led to approximately 159,260 deaths in the US in 2014 and is
the most common cancer worldwide. The increasing
relevance of pulmonary CT data has triggered dramatic growth in the
computer-aided diagnostics (CAD) field. Specifically, the CAD task for
interpreting chest CT scans can be broken down into separate steps:
delineating the lungs, detecting and segmenting nodules, and using the image
observations to infer clinical judgments. Multiple techniques
have been proposed and subsequently studied for each step. This work
focuses on characterizing the segmented nodules.

Clinical protocols for identifying and assessing nodules, specifically
the Fleischner Society Guidelines, involve monitoring the size of the
nodule with repeated scans over a period of three months to two
years. Ratings on several image-based features may also be considered,
including growth rate, spiculation, sphericity, texture, etc. Features
like size can be quantitatively estimated via image
segmentation, while other markers are mostly judged qualitatively and
subjectively. For nodule classification, existing CAD approaches are
often based on sub-optimal stratification of nodules solely based on their
morphology. Malignancy is then roughly correlated with broad morphological categories. For instance, one study found malignancy in 82\% of lobulated
nodules, 97\% of densely spiculated nodules, 93\% of ragged nodules,
100\% of halo nodules, and 34\% of round nodules~\cite{Furuya}.
Subsequent approaches incorporated automatic or manual definitions of
similar shape features, along with various other contextual or
appearance features into linear discriminant classifiers. However,
these features are mostly subjective and arbitrarily-defined
\cite{El-Baz_1}. These limitations reflect the challenges in achieving a complete and
quantitative description of malignant nodule appearances. Similarly,
it is difficult to model the 3D shape of a nodule, which is not
directly comprehensible with the routine slice-wise inspection of
human observers. Therefore, the extraction of proper appearance
features, as well as shape description, are of great value for the development of CAD
systems.

For 3D shape modeling, spherical harmonic (SH) parameterizations offer an effective model of 3-D shapes. As shape
descriptors, they have been used successfully in many applications such
as protein structure~\cite{Venkatraman}, cardiac surface
matching~\cite{Huang}, and brain mapping~\cite{Gu_1}. While SH has
been shown to successfully discriminate between malignant and benign
nodules (with 93\% accuracy for binary separation)~\cite{El-Baz_2}, using the SH coefficients to uniquely
describe a nodule's ``fingerprint'' remains largely
unexplored~\cite{El-Baz_1}. Also, as a scale- and rotation-invariant
descriptor of a mesh surface, SH dose not have the capability of
describing a nodule's size and other critical appearance features: \eg solid, sub-solid, part-solid,
peri-fissural etc. Hence, SH alone may not be sufficient for nodule
characterization.

Recently, deep convolutional neural networks (DCNNs) have been shown to
be effective at extracting image features for successful
classification across a variety of situations~\cite{ImageNet, CNNLung}. More
importantly, studies on ``transfer learning'' and using DCNN as a generic image representation~\cite{CNNTransfer,
  CNNFeature,CNNOff} have shown that successful appearance feature extraction
can be achieved without the need of significant modifications to DCNN structures, or even training on the specific dataset~\cite{CNNFeature}. While
simpler neural networks have been used for nodule
appearance~\cite{El-Baz_1}, and DCNN has recently been used to classify peri-fissural nodules~\cite{NoduleCNN}, to our knowledge, DCNNs such as the
Imagenet DCNN introduced by Krizhevsky \etal ~\cite{ImageNet} have not
been applied to the nodule malignancy problem, nor have they been
combined with 3D shape descriptors such as the SH method.

In this paper, we present a classification approach for malignancy
evaluation of lung nodules by combining both shape and appearance
features using SHs and DCNNs, respectively, on a large annotated dataset from the
Lung Image Database Consortium (LIDC)~\cite{LIDC}. First, a surface
parameterization scheme based on SH conformal mapping is used to model the variations of 3D nodule
shape. Then, a trained DCNN is used to extract the texture and intensity
features from local image patches.  Finally, the sets of DCNN and SH coefficients are combined
and used to train a random forest (RF) classifier for evaluation of their
corresponding malignancy scores, on a scale of 1 to 5. The proposed algorithm aims to
achieve a more complete description of local nodules from both shape (SH)
and appearance (DCNN) perspective. In the
following sections, we discuss the proposed method in more detail.

\section{Methods}
\label{sec:methods}
Our method works from two inputs: radiologists' binary nodule
segmentations and the local CT image patches. First, we produce a mesh
representation of each nodule from the binary segmentation using the
method from~\cite{Gu_1}. These are then mapped to the canonical
parameter domain of SH functions via conformal mapping, giving us a
vector of function coefficients as a representation of the nodule
shape. Second, using local CT images, three orthogonal local patches
containing each nodule are combined as one image input for the DCNN, and
appearance features are extracted from the first
fully-connected layer of the network. This approach for appearance
feature extraction is based on recent work in ``transfer
learning''~\cite{CNNTransfer, CNNFeature}. Finally, we combine shape
and appearance features together and use a RF classifier to assess
nodule malignancy rating.

\subsection{Spherical Harmonics Computation:}
SHs are a series of basis for representing functions defined over the
unit sphere $S^2$. The basic idea of SH parameterization is to transform a 3D
shape defined in Euclidean space into the space of SHs. In order to do
this, a shape must be first mapped onto a unit sphere. Conformal
mapping is used for this task. It functions by performing a set of
one-to-one surface transformations preserving local angles, and is
especially useful for surfaces with significant variations, such as
brain cortical surfaces~\cite{Gu_1}. Specifically, let $M$ and $N$ be
two Riemannian manifolds, then a mapping $\phi : M \to N$ will be
considered conformal if local angles between curves remain
invariant. Following the Riemann mapping theorem, a simple
surface can always be mapped to the unit sphere $S^2$, producing a
spherical parameterization of the surface.

For genus zero closed surfaces, conformal mapping is equivalent to a
harmonic mapping satisfying the  Laplace equation, $\Delta f = 0$. For
our application, nodules have approximately spherical shape, with
bounded local variations. Therefore, it is an ideal choice to use
spherical conformal mapping to normalize and parameterize the nodule
surface to a unit sphere. We first convert the binary segmentations to
meshes, and then perform conformal spherical mapping with harmonic
energy minimization. Further technical details can be found
in~\cite{Gu_1}.

With spherical conformal mapping, we are able to model the variations
of different nodule shapes onto a unit sphere. However, it is still
challenging to judge and quantify the differences within $S^2$
space. Therefore,  SHs are used to map $S^2$ to real space $\mathbb{R}$.

Similar to Fourier series as basis for the circle, SHs are capable
of decomposing a given function $f \in S^2$ into a direct sum of
irreducible sub-representations
$$
f = \sum_{l\geq0}\sum_{|m|\leq l}{\hat{f}(l,m)Y_l^m},
$$
\noindent where $Y_l^m$ is the $m$-th harmonic basis of degree $l$,
and $\hat{f}(l,m)$ is the corresponding SH coefficient. Compared to
directly using the surface in $S^2$, this gives us two major benefits:
first, the extracted representation features are rotation, scale, and
transformation invariant~\cite{Gu_1}; second, it is much easier to
compute the correlation between two vectors than two
surfaces. Therefore, SHs are a powerful representation for further
shape analysis.

\begin{figure*}[bht]
\centering
\centerline{\includegraphics[width=10cm]{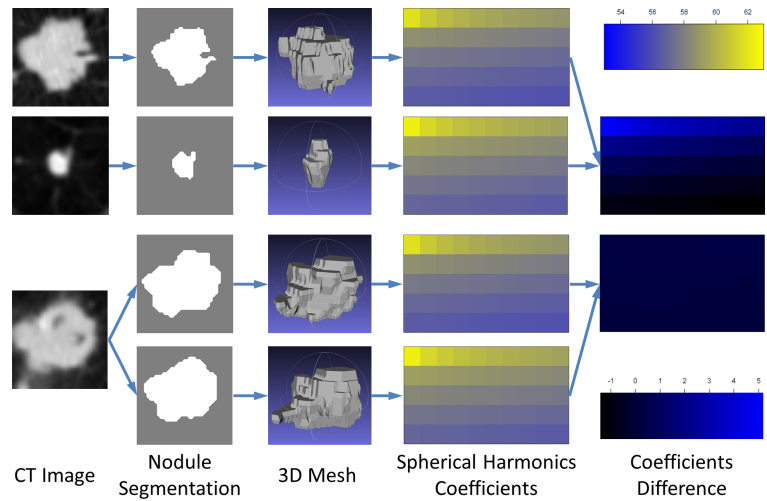}}
\caption{\small Example of SH coefficients' difference for different nodules and segmentations. The top two rows show a comparison of high-malignancy and low-malignancy nodules, and the difference of their SH coefficient values. The bottom two rows show that two different segmentations of the same nodule have much similar SH coefficients. Nonetheless, differences still remain, motivating supplementing shape-based descriptors with appearance-based ones.}
\label{fig:illu}
\end{figure*}

Fig.~\ref{fig:illu} illustrates the process of computing SH representations. It also compares the SH coefficients of four nodule segmentation cases:
nodules with high and low malignancy, and two segmentations of the
same nodule by different radiologists. From the manual segmentations,
we first generate their corresponding 3D mesh. Then, the mesh is
conformally mapped to the unit sphere and subsequently decomposed into
a series of SH coefficients. Here, we briefly compare the two
resulting SHs by using their direct difference. For comparison, the
last two rows show the SH computation for the same nodule, but with
different segmentations from two annotators. As illustrated, the SH
coefficients have far greater differences between malignant and benign
nodules than two segmentations for the same nodule, showing that it is
possible to use SH coefficients to estimate the malignancy rating of a
specific nodule. Even so, as the figure demonstrates, for nodules consisting of only a limited
number of voxels, a change in segmentation could lead to some
discrepancy in SH coefficients. For such cases, SH may not be able to serve as a
reliable marker for malignancy, and we need to assist the
classification with further information, \ie appearance.

\subsection{DCNN Appearance Feature Extraction:}
The goal of the DCNN appearance feature extraction is to obtain a
representation of local nodule patches and relate them to
malignancy. Here, we have used the same DCNN structure used by
Krizhevsky \etal~\cite{ImageNet}, which has demonstrated success in
multiple applications. This network balances discriminative power with
computational efficiency by using five convolutional layers followed
by three fully-connected layers. With the trained DCNN, each
layer provides different levels of image representations.

\begin{figure*}[bht]
\centering
\centerline{\includegraphics[width = 10cm]{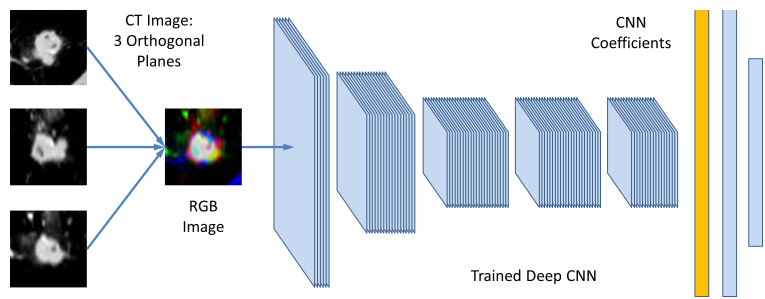}}
\caption{\small Process of appearance feature extraction. Local
  patches centered at each nodule were first extracted on three
  orthogonal planes. Then, an RGB image is generated with the three patches
  fed to each channel. This image is further resampled and used as input to a
  trained DCNN. The resulting coefficients in the first fully-connected
layer (yellow) are then used as the feature vector for nodule appearance.}
\label{fig:illuCNN}
\end{figure*}

Fig.~\ref{fig:illuCNN} shows the process how each candidate
was quantitatively coded. We first convert a local 3D CT image
volume to an RGB image, which is the required input to the DCNN structure we use
~\cite{ImageNet}. Here, we used a fixed-size cubic ROI centered at
each segmentation's center of mass with the size of the largest nodule. Since voxels in the LIDC dataset are mostly anisotropic,
we used interpolation to achieve isotropic resampling, avoiding distortion effects
in the resulting patches. In order to best preserve the appearance information, we
performed principal component analysis (PCA) on the binary
segmentation data to identify the three orthogonal axes $x', y', z'$ of the local
nodule within the regular $x, y, z$ space of axial, coronal, and sagittal
planes. Then, we resampled the local space within the
$x'y', x'z'$ and $y'z'$ planes to obtain local patches containing the
nodule. The three orthogonal patch samples formed an ``RGB'' image used as input to the DCNN's expected
three channels. We use Krizhevsky \etal{}'s pre-trained model for natural images and extract
the coefficients of the last few layers of the DCNN as a high-order representation of the input image. This ``transfer
learning'' approach from natural images has proven successful within medical-imaging domains~\cite{CNNTransfer, CNNFeature}. As an added benefit, no training of the DCNN is required, avoiding this time-consuming and computationally expensive step. For our
application, we use the first
fully-connected layer as the appearance descriptor.

\subsection{RF Classification}
By using SH and DCNN, both appearance and shape features of nodules
can be extracted as a vector of scalars, which in turn can be used together to distinguish nodules with
different malignancy ratings. Combining these two very different feature types is not trivial. Yet, recent work~\cite{CNNSVM}
has demonstrated that non-image information can be successfully combined with CNN features using classifiers. This success motivates our use
of the RF classifier to synthesize the SH and DCNN features together. The RF method features high
accuracy and efficiency, and is well-suited for problems of this
form~\cite{Brieman}. It works by ``bagging'' the data to generate new
training subsets with limited features, which are in turn used to
create a set of decision trees. A sample is then put through all trees
and voted on for correct classification. While the RF is
generally insensitive to parameter changes, we found that a set of 200
trees delivered accurate and timely performance.

\section{Experiments and Results}
\label{sec:results}
We trained and tested our method on the Lung Image Database Consortium
(LIDC) image collection~\cite{LIDC}, which consists of 1018
helical thoracic CT scans. Each scan was processed by four blinded radiologists,
who provided segmentations, shape and texture characteristic
descriptors, and also malignancy ratings. Inclusion criteria consisted of
scans with a collimation and reconstruction interval less than or
equal to 3 mm, and those with between approximately 1 and 6 lung
nodules with longest dimensions between 3 and 30 mm. The LIDC dataset was chosen for its high-quality and numerous multi-radiologist assessments.

In total 2054 nodules were extracted with 5155
segmentations, and 1432 nodules were marked by at least 2 annotators. Different
segmentations/malignancy ratings were treated individually. In order
to avoid training and testing against different segmentations of the
same nodule, dataset was split at nodule level to avoid
bias. Different segmentations of same nodules were grouped into sets based on the mean Euclidean distance between their ROI centers
using a threshold of 5 mm. To account for mis-meshing
and artifacts from interpolating slices, meshes were processed by
filters to remove holes and fill islands. We also applied 1-step
Laplacian smoothing. 

Judging from the distribution of malignancy ratings for all annotating
radiologists and based on Welch's \textit{t}-test, inter-observer
differences is significant among annotators. Meanwhile, according to the range of malignancy rating differences for any specific nodule, most nodules have a rating discrepancy of 2 or 3 among different annotators, indicating that inter-observer variability is highly significant. Therefore to evaluate the performance of the proposed framework, we used ``off-by-one'' accuracy, meaning that we regard a malignancy rating with $\pm 1$ as a reasonable and acceptable evaluation.

Accuracy results for 10-fold cross validation are shown in Table~\ref{table:result} for a range of nodule sets and SH
coefficients. Three sets of models were used, one using DCNN features
only, one using SH coefficients only, and one using both SH and DCNN
features. Models were tested with a range of input parameters, including maximum number of
coefficients included and minimum number of annotators marking  the
nodule. In all cases, the hybrid model achieved better results than
both individual models using the same input parameters. The hybrid model results are even more impressive when compared against the inter-observer variability of the LIDC dataset. These results indicate
that DCNNs and SHs provide complementary appearance and feature
information that can help providing reference malignancy ratings of lung nodules.

\begin{table}[t]
\vspace*{-5pt}
\centering
\caption{\small Off-by-one accuracy for SH only, DCNN only, and
  hybrid models for input sets of number of annotators marking the
  nodule, and maximum number of SH coefficients included.}
\label{table:result}
\setlength{\tabcolsep}{3pt}
\begin{tabular}{c|c|c|c|c}
Min Annotations &  \# of SH Coeffs & DCNN only & SH only & SH+DCNN \\
\hline
1 & 100 & \multirow{3}{*}{0.791}& 0.772 & \textbf{0.812}\\
1 & 150 &  & 0.783 & \textbf{0.824}\\
1 & 400 &  & 0.774 & \textbf{0.807}\\
\hline
2 & 100 & \multirow{3}{*}{0.759} & 0.761 & \textbf{0.803}\\
2 & 150 & & 0.779 & \textbf{0.793}\\
2 & 400 & & 0.761 & \textbf{0.824}\\
\end{tabular}
\end{table}

\section{Discussion and Conclusion}
In this study, we presented an approach for generating a
reference opinion about lung nodule
malignancy based on the knowledge of experts' characterizations. Our method is based on hybrid feature
sets that include shape
features, from SHs decomposition, and appearance features, from a DCNN trained on natural images. Both features are
subsequently used for malignancy classification with a RF
classifier.

There are many promising avenues of future work. For instance, the method would benefit even more from a larger and more accurate
testing pool, as well as the inclusion of more reliable and precise
ground truth data beyond experts' subjective evaluations. In addition, using additional complementary
information, such as volume and scale-based features, may also further
improve scores. In this study, we represented a nodule's appearance within orthogonal planes along three PCA
axis. Indeed, including more 2D views, even 3D DCNN, could potentially
be meaningful beyond the promising results from current setting. The rating
classification can also be formulated as regression, whereas the
results were not statistically significant according to our current
experiment.

SH computation variations due to nodule size and segmentation remains open and discussion is limited in existing
literatures [6]. In this study, our experiments 
partially covered this robustness via testing segmentations for same
nodules from different human observers. We also observed that including more SH coefficients did not necessarily
led to higher accuracy. We postulate that
coefficients help define shape to a certain point, beyond which it may
introduce more noise than useful information, and further
investigation would be helpful to test this hypothesis.

Based on the inter-observer variability, experimental results using the
LIDC dataset demonstrate that the proposed scheme can perform comparably to an independent expert annotator, but does so using full
automation up to segmentation. As a result, this work serves as an important demonstration of how both shape and appearance information can be harnessed for the important task of lung nodule classification.

\bibliographystyle{splncs}
\bibliography{refs}

\begin{thebibliography}{10}

\bibitem{Furuya}
Furuya, K., Murayama, S., Soeda, H., Murakami, J., Ichinose, Y., Yauuchi, H.,
  Katsuda, Y., Koga, M., Masuda, K.:
\newblock New classification of small pulmonary nodules by margin
  characteristics on highresolution {CT}.
\newblock Acta Radiologica \textbf{40} (1999)  496--504

\bibitem{El-Baz_1}
El-Baz, A., Beache, G.M., Gimel'farb, G., Suzuki, K., Okada, K., Elnakib, A.,
  Soliman, A., Abdollahi, B.:
\newblock Computer-aided diagnosis systems for lung cancer: Challenges and
  methodologies.
\newblock International Journal of Biomedical Imaging \textbf{2013} (2013)
  942353

\bibitem{Venkatraman}
Venkatraman, V., Sael, L., Kihara, D.:
\newblock Potential for protein surface shape analysis using spherical
  harmonics and {3D} {Zernike} descriptors.
\newblock Cell Biochemistry and Biophysics \textbf{54}(1-3) (2009)  23--32

\bibitem{Huang}
Huang, H., Shen, L., Zhang, R., Makedon, F., Hettleman, B., Pearlman, J.:
\newblock Surface alignment of {3D} spherical harmonic models: Application to
  cardiac {MRI} analysis.
\newblock In: Medical Image Computing and Computer-Assisted Intervention –
  MICCAI 2005. Volume 3749 of Lecture Notes in Computer Science.
\newblock Springer Berlin Heidelberg (2005)  67--74

\bibitem{Gu_1}
Gu, X., Wang, Y., Chan, T.F., Thompson, P.M., tung Yau, S.:
\newblock Genus zero surface conformal mapping and its application to brain
  surface mapping.
\newblock IEEE Transactions on Medical Imaging \textbf{23} (2004)  949--958

\bibitem{El-Baz_2}
El-Baz, A., Nitzken, M., Khalifa, F., Elnakib, A., Gimel’farb, G., Falk, R.,
  El-Ghar, M.:
\newblock 3d shape analysis for early diagnosis of malignant lung nodules.
\newblock In: Information Processing in Medical Imaging. Volume 6801 of Lecture
  Notes in Computer Science.
\newblock Springer Berlin Heidelberg (2011)  772--783

\bibitem{ImageNet}
Krizhevsky, A., Sutskever, I., Hinton, G.E.:
\newblock Imagenet classification with deep convolutional neural networks.
\newblock In Pereira, F., Burges, C.J.C., Bottou, L., Weinberger, K.Q., eds.:
  Advances in Neural Information Processing Systems 25.
\newblock Curran Associates, Inc. (2012)  1097--1105

\bibitem{CNNLung}
Gao, M., Bagci, U., Lu, L., Wu, A., Buty, M., Shin, H.C., Roth, H., Papadakis,
  G.Z., Depeursinge, A., Summers, R., Xu, Z., Mollura, D.J.:
\newblock Holistic classification of ct attenuation patterns for interstitial
  lung diseases via deep convolutional neural networks.
\newblock In: 1st Workshop on Deep Learning in Medical Image Analysis. DLMIA
  2015 (October 2015)  41--48

\bibitem{CNNTransfer}
Shin, H.C., Roth, H.R., Gao, M., Lu, L., Xu, Z., Nogues, I., Yao, J., Mollura,
  D., Summers, R.M.:
\newblock Deep convolutional neural networks for computer-aided detection: Cnn
  architectures, dataset characteristics and transfer learning.
\newblock IEEE Transactions on Medical Imaging \textbf{PP}(99) (2016)  1--1

\bibitem{CNNFeature}
Bar, Y., Diamant, I., Wolf, L., Lieberman, S., Konen, E., Greenspan, H.:
\newblock Chest pathology detection using deep learning with non-medical
  training.
\newblock In: Biomedical Imaging (ISBI), 2015 IEEE 12th International Symposium
  on. (April 2015)  294--297

\bibitem{CNNOff}
Razavian, A., Azizpour, H., Sullivan, J., Carlsson, S.:
\newblock Cnn features off-the-shelf: an astounding baseline for recognition.
\newblock In: Proceedings of the IEEE Conference on Computer Vision and Pattern
  Recognition Workshops. (2014)  806--813

\bibitem{NoduleCNN}
Ciompi, F., de~Hoop, B., van Riel, S.J., Chung, K., Scholten, E.T., Oudkerk,
  M., de~Jong, P.A., Prokop, M., van Ginneken, B.:
\newblock Automatic classification of pulmonary peri-fissural nodules in
  computed tomography using an ensemble of 2d views and a convolutional neural
  network out-of-the-box.
\newblock Medical Image Analysis \textbf{26}(1) (2015)  195 -- 202

\bibitem{LIDC}
Armato, S.G., McLennan, G., Bidaut, L., et~al.:
\newblock The lung image database consortium ({LIDC}) and image database
  resource initiative ({IDRI}): A completed reference database of lung nodules
  on {CT} scans.
\newblock Medical Physics \textbf{38}(2) (2011)  915--931

\bibitem{CNNSVM}
Sampaio, W.B., Diniz, E.M., Silva, A.C., de~Paiva, A.C., Gattass, M.:
\newblock Detection of masses in mammogram images using cnn, geostatistic
  functions and \{SVM\}.
\newblock Computers in Biology and Medicine \textbf{41}(8) (2011)  653 -- 664

\bibitem{Brieman}
Breiman, L.:
\newblock Random forests.
\newblock Machine Learning \textbf{45}(1) (2001)  5--32

\end{thebibliography}
\end{document}